\documentclass{article}

\usepackage{microtype}
\usepackage{graphicx}
\usepackage{titling} 
\usepackage{subfigure}
\usepackage{amsmath} 
\usepackage{booktabs} 
\usepackage{multirow}
\usepackage{mathrsfs}
\usepackage{algpseudocode}
\usepackage{colortbl}
\usepackage{xcolor}
\usepackage{enumitem}
\usepackage{float}  
\usepackage{hyperref}

\usepackage[accepted]{icml2025}
\usepackage{amssymb}
\usepackage{mathtools}
\usepackage{amsthm}
\usepackage{dsfont}

\usepackage[capitalize,noabbrev]{cleveref}

\theoremstyle{plain}

\theoremstyle{definition}

\theoremstyle{remark}

\usepackage[textsize=tiny]{todonotes}

\icmltitlerunning{Submission and Formatting Instructions for ICML 2025}

\begin{document}

\twocolumn[
\icmltitle{Sce2DriveX: A Generalized MLLM Framework for Scene-to-Drive Learning}



\icmlsetsymbol{equal}{*}

\begin{icmlauthorlist}
\icmlauthor{Rui Zhao}{yyy,comp}
\icmlauthor{Qirui Yuan}{yyy}
\icmlauthor{Jinyu Li}{yyy}
\icmlauthor{Haofeng Hu}{yyy}
\icmlauthor{Yun Li}{sch}
\icmlauthor{Chengyuan Zheng}{yyy}
\icmlauthor{Fei Gao}{yyy,comp}
\end{icmlauthorlist}

\icmlaffiliation{yyy}{Department of XXX, University of YYY, Location, Country}
\icmlaffiliation{comp}{Company Name, Location, Country}
\icmlaffiliation{sch}{School of ZZZ, Institute of WWW, Location, Country}

\icmlcorrespondingauthor{Firstname1 Lastname1}{first1.last1@xxx.edu}
\icmlcorrespondingauthor{Firstname2 Lastname2}{first2.last2@www.uk}

\icmlkeywords{Machine Learning, ICML}

\vskip 0.3in
]




\begin{abstract}
End-to-end autonomous driving, which directly maps raw sensor inputs to low-level vehicle controls, is an important part of Embodied AI. Despite successes in applying Multimodal Large Language Models (MLLMs) for high-level traffic scene semantic understanding, it remains challenging to effectively translate these conceptual semantics understandings into low-level motion control commands and achieve generalization and consensus in cross-scene driving. We introduce Sce2DriveX, a human-like driving chain-of-thought (CoT) reasoning MLLM framework. Sce2DriveX utilizes multimodal joint learning from local scene videos and global BEV maps to deeply understand long-range spatiotemporal relationships and road topology, enhancing its comprehensive perception and reasoning capabilities in 3D dynamic/static scenes and achieving driving generalization across scenes. Building on this, it reconstructs the implicit cognitive chain inherent in human driving, covering scene understanding, meta-action reasoning, behavior interpretation analysis, motion planning and control, thereby further bridging the gap between autonomous driving and human thought processes. To elevate model performance, we have developed the first extensive Visual Question Answering (VQA) driving instruction dataset tailored for 3D spatial understanding and long-axis task reasoning. Extensive experiments demonstrate that Sce2DriveX achieves state-of-the-art performance from scene understanding to end-to-end driving, as well as robust generalization on the CARLA Bench2Drive benchmark.
\end{abstract}

\begin{figure*}[!t]
\centering
\includegraphics[width=1\textwidth]{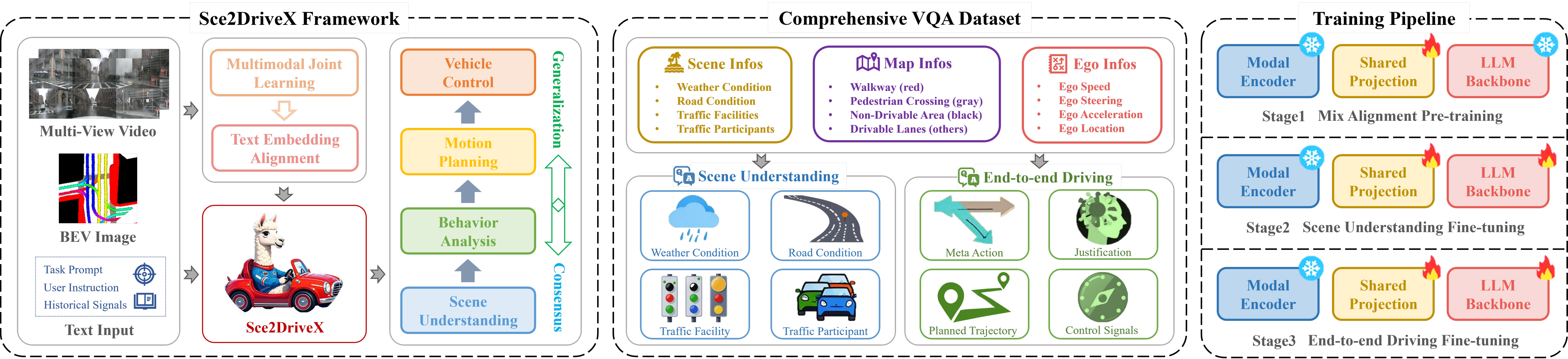} 
\caption{\textbf{Overall pipeline.} Our goal is to achieve generalization and consensus in cross-scene driving, including three novel contributions: 1) Sce2DriveX framework; 2) comprehensive VQA driving instruction dataset; and 3) three-stage training pipeline.}
\label{fig1}
\vspace{-5 pt} 
\end{figure*}

\section{Introduction}
\label{submission}

Embodied AI equips intelligent agents like autonomous driving (AD) models with the ability to perceive, reason, and interact with the real world in real-time. However, a core challenge lies in the generalization and consensus of AD model frameworks.On one hand, AD learning frameworks may struggle with generalizing complex, dynamic traffic scenes like varied weather conditions, road layouts, traffic semantics, and the behavioral preferences of surrounding participants. On the other hand, AD systems' decision-making strategies often lack alignment with human drivers' cognitive processes, complicating humans' understanding of system behavior. These challenges arise from the gap between high-level scene semantic understanding and low-level motion control commands. Developing a human-like framework capable of all-weather, all-scene perception and reasoning has thus become a widely discussed topic.

Current AD research often employs small-model learning frameworks \cite{zeng2019end,hu2022st,hu2023planning}. Due to the limited reasoning capabilities of small models, these systems produce rigid responses to predefined problems, making it difficult to deliver satisfactory results when faced with new or unexpected queries. 

Recently, the rapid development of MLLMs \cite{li2023blip,liu2024improved,driess2023palm} has demonstrated significant advantages in various vision-language tasks. By leveraging MLLMs as a bridge between high-level scene semantic understanding and low-level motion control commands, we can address the challenges of generalization and consensus in AD models. Benefiting from pretraining on extensive cross-modal and cross-disciplinary data, MLLMs offer strong reasoning and generalization capabilities, enabling them to manage diverse scenarios and enhance adaptable cross-scene driving. Additionally, MLLMs' robust text query and cognitive abilities allow them to align driving thoughts with human consensus and translate complex reasoning into understandable natural language, providing a unified explanatory layer for AD. However, AD is a complex task characterized by spatiotemporal continuity, dynamic scenarios, and global coordination. Current MLLM-based AD research primarily uses single-frame front-view scene images \cite{sima2025drivelm} as perceptual input, leading to a lack of deep understanding of spatiotemporal relationships and road features, along with inadequate traffic scene comprehension. Moreover, when generating driving commands, current studies often only map scene factors to low-level control signals \cite{xu2024drivegpt4}, overlooking the reasoning behind future vehicle behaviors, failing to utilize MLLMs' capabilities for generalized cognitive reasoning and diverging from human driving thought. 

In addition to the model framework, matched datasets are crucial for efficient training and performance ceiling of the model. Many datasets are designed in the form of VQA, although some success has been achieved, models trained on existing VQA datasets still face limitations in addressing the complexities of AD. This limitation primarily stems from the disparity in visual information between traffic scenes and VQA datasets [9, 10], requiring models to effectively utilize the complementary information of multimodal perceptual data to understand complex scenes and capture object dynamics from multi-frame data streams. Furthermore, most VQA datasets are tailored to single driving tasks. They often provide only simple Boolean answers (i.e., yes or no) or limited multiple-choice responses \cite{qian2024nuscenes} in closed-ended question annotations, lacking in richness.

To address these gaps, this paper proposes Sce2DriveX framework (shown on the left side of Figure \ref{fig1}), which uses modal encoders to align visual representations of multi-view scene videos and BEV map images into a unified visual feature space, which is then mapped to a text embedding space through a shared projection and processed by the LLM backbone to generate natural language responses including scene understanding, behavior analysis, motion planning, and vehicle control. This multimodal joint learning of local scenes and global maps endows model a deep understanding of spatiotemporal relationships and road topology, extending its ability for comprehensive 3D dynamic/static scene perception and reasoning, thereby achieving generalization and consensus driving across scenes. To support training, this paper constructs the first comprehensive VQA driving instruction dataset (shown in the middle of Figure \ref{fig1}) for 3D spatial understanding and long-axis task reasoning, focusing on hierarchical scene understanding and interpretable end-to-end driving tasks in multimodal, multi-view, and multi-frame contexts. Additionally, it details a task-oriented three-stage training pipeline for supervised fine-tuning (shown on the right side of Figure \ref{fig1}), including mix alignment pre-training, scene understanding fine-tuning, and end-to-end driving fine-tuning. The key contributions of this paper are as follows:

\begin{itemize}[left=0pt, itemsep=0pt]
\item We propose Sce2DriveX, a human-like CoT reasoning MLLM framework, aimed to achieve progressive reasoning learning from multi-view long-range scene understanding to behavior analysis, motion planning, and vehicle control driving process.

\item We construct the first comprehensive VQA driving instruction dataset for 3D spatial understanding and long-axis task reasoning, and introduce a task-oriented three-stage training pipeline to enhance the perception-reasoning ability of Sce2DriveX.

\item Extensive experiments demonstrate that Sce2DriveX achieves state-of-the-art performance across tasks including scene understanding, meta-action reasoning, behavior interpretation analysis, motion planning, and control signal generation.
\end{itemize}

\begin{figure*}[!t]
\centering
\includegraphics[width=0.8\textwidth]{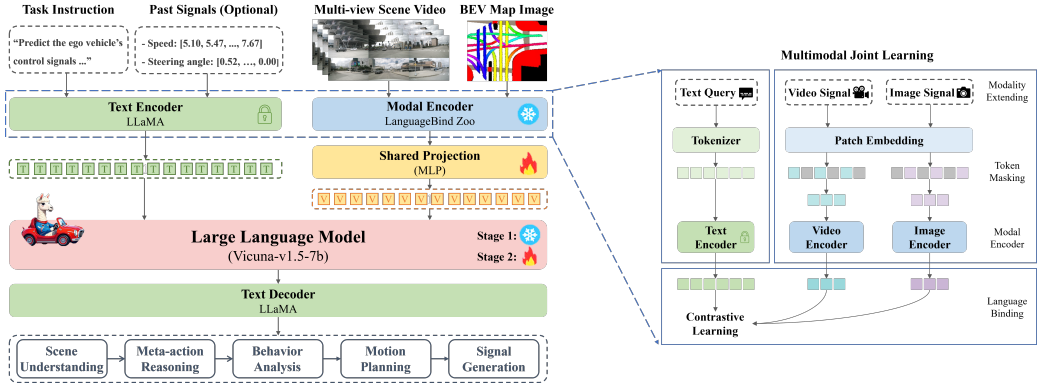} 
\caption{\textbf{Model architecture.} Sce2DriveX uses modal encoders to emergently align the visual representations of multi-view scene videos and BEV map images into a unified visual feature space, which are then mapped to the LLM backbone through a shared projection.}
\label{fig2}
\vspace{-5pt} 
\end{figure*}

\section{Related Works}
\label{submission}

\subsection{Multimodal Large Language Models}

Recent advancements have seen the creation of MLLMs. Flamingo \cite{alayrac2022flamingo} and BLIP2 \cite{li2023blip} align visual features with the embedding space of LLM through gated attention and Q-Former, while LLaVA \cite{liu2024improved} and MiniGPT4 \cite{zhu2023minigpt} use Multilayer Perceptrons (MLPs) to combine pretrained vision module with LLM backbone. Additionally, some studies have attempted to extend modality interaction to video and audio. Video-LLaVA \cite{lin2023video} employs a LanguageBind encoder to pre-align different visual features to the text space, facilitating joint training for images and videos. Video-Llama \cite{zhang2023video} achieves joint processing of visual and auditory signals in video data by integrating pretrained visual and audio encoder into the LLM.

\subsection{Autonomous Driving With Multimodal Large Language Models}

MLLMs have demonstrated the potential to understand traffic scenes, optimize driving decisions, and fundamentally improve human-vehicle interactions. Compared to traditional AD perception systems, MLLMs offer a new paradigm, leveraging their inherent few-shot learning capabilities to rapidly learn from vast amounts of multimodal data, thereby providing richer supervision sources. PromptTrack \cite{wu2023language} integrates cross-modal features into language prompts as semantic cues, combined with MLLM for 3D detection and tracking.  Talk2BEV \cite{choudhary2023talk2bev} combines BEV images with language prompts, using MLLM for audiovisual integration in AD. For end-to-end driving, MLLMs also exhibit better interpretability and trustworthiness. DriveGPT4 \cite{xu2024drivegpt4} pioneers using MLLM to transform sensor data and instructions into control signals and text responses. RAG-Driver \cite{yuan2024rag} proposes a retrieval-augmented MLLM that generates driving behavior justifications and predicts control signals by retrieving expert demonstrations. DriveVLM \cite{tian2024drivevlm} integrates the cognitive chain module into the MLLM, enabling driving scene description and motion planning. However, Existing research has yet to align MLLMs with the implicit cognitive chain of human driving, enabling reasoning from combined global and local scene understanding to behavior, trajectory, and control commands, limiting cross-scene generalization and human-consensus driving.

\subsection{Visual Question Answering Datasets}

To support the efficient training of MLLMs, the design of large-scale VQA datasets has become a research hotspot. Currently, various VQA datasets exist, including image-based datasets such as CLEVR \cite{johnson2017clevr}, VQA2.0 \cite{goyal2017making}, and EQA \cite{das2018embodied}, as well as video-based datasets like TVQA \cite{lei2018tvqa}, TGIF-QA \cite{jang2017tgif}, and ActivityNet-QA \cite{yu2019activitynet}. For ImageQA, early studies \cite{johnson2017clevr,fukui2016multimodal} attempted to fuse image features extracted by Convolutional Neural Networks (CNNs) with question encodings, which were then fed into decoders for answer generation. Recently, Transformer-based models \cite{tan2019lxmert,zhang2021vinvl} have achieved state-of-the-art performance in ImageQA tasks. Through attention networks, some studies have effectively captured the intrinsic relationships between temporal context and spatial features in video frames. 3D QA is a novel task in the VQA domain, focusing on answering questions about 3D-view scenes, requiring models to understand the geometric structures and spatial relationships of objects. Recently, many 3D QA datasets have been constructed, such as 3DQA \cite{ye2022visatlas}, ScanQA \cite{azuma2022scanqa}, and SQA3D \cite{ma2022sqa3d}. Despite significant progress in the VQA community, challenges remain when dealing with complex traffic scenes involving multimodal, multi-view, and multi-frame contexts. Moreover, the AD field currently lacks comprehensive VQA driving datasets.

\section{Methodology}
\label{submission}

\subsection{Sce2DriveX Framework}

This paper aims to develop a human-like CoT reasoning MLLM framework, enabling progressive reasoning learning from multi-view long-range scene understanding to behavior analysis, motion planning, and vehicle control driving process. As shown in Figure \ref{fig2}, See2DriveX consists of four components: 1) Modal Encoder, including a video encoder \( f_{\mathrm{V\_E}} \) and an image encoder \( f_{\mathrm{I\_E}} \), initialized from OpenCLIP; 2) Shared Projection \( f_{\mathrm{P}} \), using two fully connected layers with GeLU activation; 3) LLM Backbone \( f_{\mathrm{LLM}} \), employing Vicuna-v1.5-7b; 4) Text Encoder \( f_{\mathrm{T\_E}} \) and Text Decoder \( f_{\mathrm{T\_D}} \), provided by LLaMA.

\subsubsection{Multimodal Joint Learning}

Given text instruction $\mathrm{X}_\mathrm{T}$, we first use a Byte Pair Encoding (BPE) tokenizer to segment the words into relatively common subwords, each corresponding to a unique logit. Then, these logits are encoded using the text encoder $f_{\mathrm{T\_E}}$:

\begin{equation}
    \mathrm{H}_\mathrm{T} = f_{\mathrm{T\_E}} (BPE(\mathrm{X}_\mathrm{T}))
\end{equation}

where \( \mathrm{H}_\mathrm{T} \in \mathbb{R}^{M_{txt} \times L} \) represents the text tokens, \( M_{txt} \) is the number of  tokens, and \( L \) is the LLM's feature dimension.

Given multi-view scene video \( \mathrm{X}_\mathrm{V} \in \mathbb{R}^{T \times H \times W \times C} \) and BEV map image \( \mathrm{X}_\mathrm{I} \in \mathbb{R}^{H \times W \times C} \), where \( T \) is the number of video frames, \( (H, W) \) is the resolution of the original image, and \( C \) is the number of channels, we adopt a patch masking method. By using an encoder mask \( \mathbb{M}_{e} \), a small subset of patches is selected and segmented to alleviate the issue of excessive token numbers in the modal encoder. Specifically, the video signal \( \mathrm{X}_\mathrm{V} \) and image signal \( \mathrm{X}_\mathrm{I} \) are first converted into corresponding patches \( \mathrm{P}_\mathrm{V} \in \mathbb{R}^{T \times N_{p} \times C} \) and \( \mathrm{P}_\mathrm{I} \in \mathbb{R}^{N_{p} \times C} \) through patch embedding layer with non-overlapping filters, where \( N_{p} = \frac{H \times W}{B^{2}} \) is the number of patches, and \( B \) is the size of each patch. Then, positional embeddings are applied to the visible tokens, and they are divided using the encoder mask. The combined video sequence \( \mathrm{S}_\mathrm{V} \) and image sequence \( \mathrm{S}_\mathrm{I} \) are represented as:

\begin{equation}
    \mathrm{S}_\mathrm{V} = \{ \mathrm{P}_\mathrm{V} + \mathrm{Q}_i \}_{i \in \mathbb{M}_e}, \quad \mathrm{S}_\mathrm{I} = \{ \mathrm{P}_\mathrm{I} + \mathrm{Q}_j \}_{j \in \mathbb{M}_e}
\end{equation}

where $\mathrm{Q}$ represents a series of learnable positional tokens, and $i$ and $j$ denote the positional information of video patches and image patches, respectively.

Finally, the video sequence $\mathrm{S}_\mathrm{V}$ is encoded using the video encoder $f_{\mathrm{V\_E}}$, and the image sequence $\mathrm{S}_\mathrm{I}$ is encoded using the image encoder $f_{\mathrm{I\_E}}$:

\begin{equation}
    \mathrm{H}_\mathrm{V} = f_{\mathrm{V\_E}}(\mathrm{S}_\mathrm{V}), \quad \mathrm{H}_\mathrm{I} = f_{\mathrm{I\_E}}(\mathrm{S}_\mathrm{I})
\end{equation}

where $\mathrm{H}_\mathrm{V} \in \mathbb{R}^{T\times M_{vid} \times V}$ represents the video tokens, $\mathrm{H}_\mathrm{I} \in \mathbb{R}^{M_{img} \times V}$ represents the image tokens,$ M_{vid}$ is the number of video tokens,$ M_{img}$ is the number of image tokens, and $V$ is the unified visual feature dimension. Notably, to achieve multimodal semantic alignment, we adopt the modal encoding approach of LanguageBind \cite{zhu2023languagebind}, which uses text as the bridge between different modalities. Through contrastive learning principles, other modalities are bound to the text modality and emergently aligned to the unified visual feature space.

\subsubsection{Unified Processing by the LLM Backbone}

Our goal is to map multimodal tokens into the text embedding space, providing a unified visual representation for the LLM, which is then combined with tokenized text queries and fed into the LLM backbone to generate responses. Specifically, we first use the shared projection $f_{\mathrm{P}}$ to map the video tokens $\mathrm{H}_\mathrm{V}$ and image tokens $\mathrm{H}_\mathrm{I}$:

\begin{equation}
    \mathrm{H}_\mathrm{L} = f_{\mathrm{P}}(\mathrm{H}_\mathrm{V}, \mathrm{H}_\mathrm{I})
\end{equation}

where $\mathrm{H}_\mathrm{L} \in \mathbb{R}^{T \times M_{vsl} \times L}$ represents the unified visual tokens, which share the same feature dimension as the text tokens $\mathrm{H}_\mathrm{T}$, $M_{vsl}$ is the number of visual tokens. Next, the unified visual tokens $\mathrm{H}_\mathrm{L}$ are combined with the text tokens $\mathrm{H}_\mathrm{T}$ and fed into the LLM backbone $f_{\mathrm{LLM}}$ for processing, generating corresponding predicted tokens. These predicted tokens are finally decoded back into natural language responses $\mathrm{Z}$ by the text decoder $f_{\mathrm{T\_D}}$:

\begin{equation}
    \mathrm{Z} = f_{\mathrm{T\_D}}(f_{\mathrm{LLM}}(\mathrm{H}_\mathrm{L} \oplus \mathrm{H}_\mathrm{T}))
\end{equation}

where $\oplus$ denotes the concatenation operation, $\mathrm{Z} \in \{ \mathrm{Z}_{sce}, \mathrm{Z}_{act}, \mathrm{Z}_{int}, \mathrm{Z}_{mot}, \mathrm{Z}_{sig} \}$ includes scene understanding $\mathrm{Z}_{sce}$, meta-action reasoning $\mathrm{Z}_{act}$, behavior interpretation analysis $\mathrm{Z}_{int}$, motion planning $\mathrm{Z}_{mot}$, and control signal generation $\mathrm{Z}_{sig}$.

In summary, Sce2DriveX understands long-range spatiotemporal relationships and road topology, enhancing 3D scene perception and reasoning for cross-scene driving generalization. It also restores the cognitive chain of human driving, strengthening consensus between AD and human thought.

\begin{figure*}[!t]
\centering
\includegraphics[width=1\textwidth]{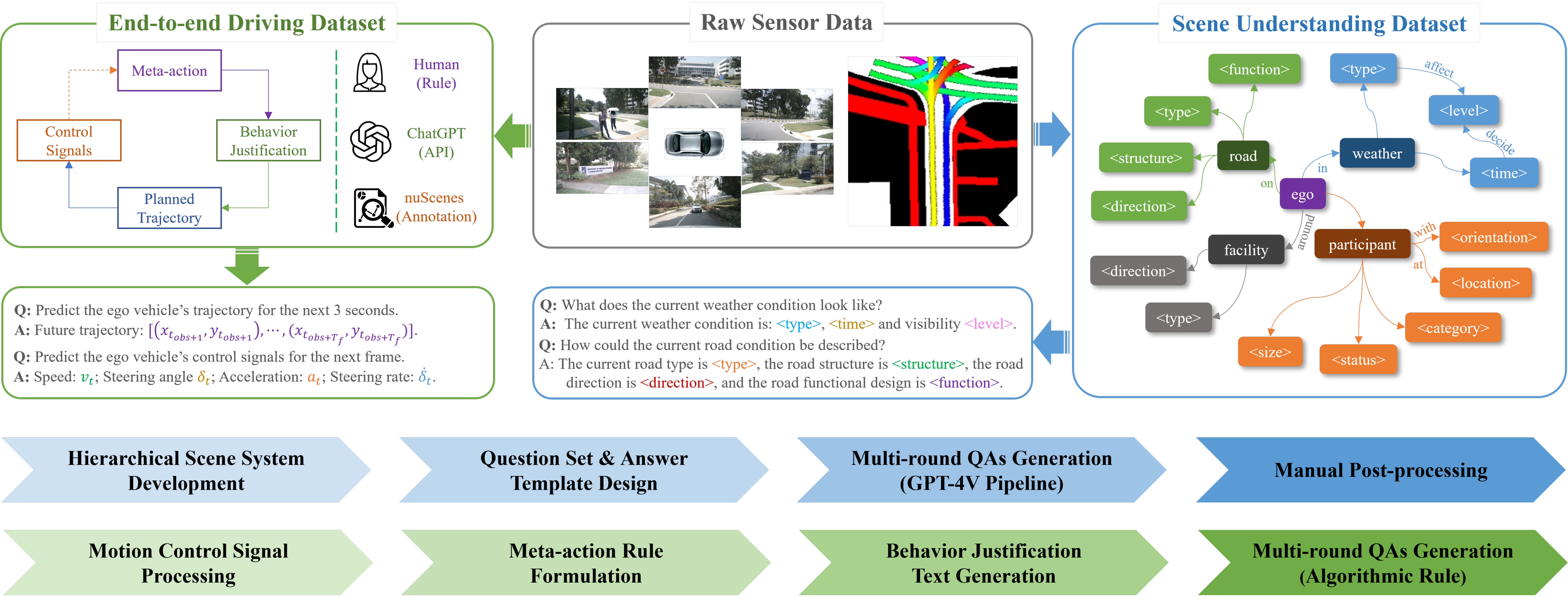} 
\caption{\textbf{Dataset construction.} Comprehensive VQA driving instruction dataset consists of two subsets: 1) Hierarchical Scene Understanding Dataset; 2) Interpretable End-to-End Driving Dataset.}
\label{fig3}
\vspace{-5pt} 
\end{figure*}

\subsection{VQA Driving Instruction Dataset}

To train Sce2DriveX, this paper constructs the first comprehensive VQA driving instruction dataset for 3D spatial understanding and long-axis task reasoning. Based on the open-source nuScenes \cite{caesar2020nuscenes}, it integrates structured data from three modalities: multi-view scene videos (local), BEV map images (global), and multi-round QA annotations. As shown in Figure \ref{fig3}, the dataset includes two subsets: 1) Hierarchical Scene Understanding Dataset; 2) Interpretable End-to-End Driving Dataset.

\subsubsection{Hierarchical Scene Understanding}

The hierarchical scene understanding dataset is generated through a scalable automated pipeline, providing a hierarchical, structured description of traffic scenes. It covers weather, roads, facilities, and traffic participants (3D), challenging the model's 3D spatial understanding. Figure \ref{fig3} shows its construction process, including hierarchical scene development, question set \& answer template design,multi-round QA generation, and manual post-processing.

\textbf{Hierarchical Scene System Development} To enhance the model's recognition of long-tail scenes, we develop a novel hierarchical scene system $E$ to hierarchically describe traffic scenes, covering four elements: $E=\{E_{weather},E_{road},E_{facility},E_{participant}\}$. Each element has multiple attributes. Specifically, weather condition $E_{weather}$ includes $<type>$, $<time>$, and $<level>$, reflecting driving difficulty. Road condition $E_{road}$ includes $<type>$, $<structure>$, $<direction>$, and $<function>$, determining driving patterns. Traffic facility $E_{facility}$ is categorized into traffic signs and road markings, regulating driving behavior. We define their inherent attributes, including $<type>$ and $<direction>$. Traffic participant $E_{participant}$ encompasses $<category>$, $<status>$, $<size>$, $<location>$, and $<orientation>$, affecting driving states. Based on the above attributes, we annotate two features for each traffic participant: 3D intrinsic features, capturing static attributes such as category, 3D dimensions, current status, location, and orientation; and 2D motion features, describing dynamic behavior through current and historical locations. Further details are provided in the appendix.

We construct a scene graph to visualize the hierarchical scene system. As shown in Figure \ref{fig3} (right), the graph enhances the visual scene with structured relations: the central node represents the ego vehicle, intermediate nodes represent the four scene elements, and outermost nodes denote their attributes. Nodes are connected by edges representing actions (verbs) or spatial relationships (prepositions). 

\textbf{Question Set \& Answer Template Design} Based on the hierarchical scene system, we manually design four question sets centered around each scene element: $Q=\{Q_{weather},Q_{road},Q_{facility},Q_{participant}\}$. To avoid overfitting to fixed question patterns, each set includes multiple synonymous expressions for a single question option. To achieve this, we consult relevant grammatical rules and collected various synonymous question patterns, ultimately creating at least two grammatically correct and semantically equivalent expressions for each question option.

Using the scene graph, we manually design four answer templates $A=\{A_{weather},A_{road},A_{facility},A_{participant}\}$ that match the question sets, incorporating various parameterized attributes. We traverse the scene graph's nodes and edges, convert them into [element-relation-attribute] triplets, and write fixed text for each. Detailed question sets and answer templates are in the appendix.

\textbf{Multi-round QA Generation and Manual Post-processing}  We use an automated pipeline with ChatGPT to generate multi-round QA annotations. This pipeline processes multi-view scene frames and BEV map images, using question sets and templates for context. ChatGPT samples questions, reasons with visual data, and outputs instantiated templates covering situational awareness, object recognition, tracking, and spatial localization.

A key challenge is hallucination, where text mismatches visuals. To mitigate this, we manually refine annotations by removing inappropriate QAs, correcting errors, and filling in missing options.

\subsubsection{Interpretable End-to-End Driving}

The interpretable end-to-end driving dataset is automatically integrated through intent-cognition-oriented algorithmic rules, achieving a sequential and transparent description of the driving process. It covers meta-action, behavior justification, planned trajectory, and control signals (multi-type), challenging the model's long-axis task reasoning capabilities. Figure 3 (left) illustrates its construction process, including motion control signal processing, meta-action rule formulation, and behavior justification text generation.

\textbf{Motion Control Signal Processing} The motion control signals $S$ include the planned trajectory $\mathcal{W} = \{ ( x_{t}, y_{t} ) \}_{t = t_{obs} - T_{h}}^{T_{f}}$ and the low-level control signals $\mathcal{U} = \{a_t, \dot{\delta}_t, v_{t+1}, \delta_{t+1}\}_{t = t_{obs} - T_h}^{t_{obs}}$, where $\dot{\delta}_{t}$ is the steering rate and $\delta_{t+1}$ is the steering angle. The original nuScenes annotations provide accompanying motion control signals for each scene. We parse these signals into structured JSON entries and classify them. Specifically, we integrate the historical trajectory and control signals as known information, with task background text, into system prompts to assist reasoning. Additionally, we use the future trajectory and current/next frame control signals as ground truth labels, filling them into predefined templates to refine predictions.

\textbf{Meta-action Rule Formulation} The ego vehicle's meta-action $\mathcal{A}$ is represented as a combination of lateral speed level estimation $\mu^{sp\_x}$, longitudinal speed level estimation $\mu^{sp\_y}$, and steering level estimation $\mu^{st}$: $\mathcal{A} := (\mu^{sp\_x} *\mu^{st}*\mu^{sp\_y})$. We define a threshold space $\Omega = [\varepsilon_{min}^{a\_x}, \varepsilon_{max}^{a\_x}, \varepsilon_{min}^{a\_y}, \varepsilon_{max}^{a\_y}, \varepsilon^{v}, \varepsilon^{\Delta x}, \varepsilon^{\Delta\theta}]$. Based on $\Omega$, we detail the judgment rules for each component. The $\mu^{sp\_x}$ and $\mu^{sp\_y}$ are determined by the relationship between the acceleration in the current frame and predefined thresholds. The $\mu^{st}$ follows a hierarchical judgment principle, first excluding idle and move straight actions based on the current frame, then traversing future timesteps that satisfy the straight-moving condition. The turn left/right and slight to left/right are determined based on the relationship between yaw rate gain, lateral displacement, and thresholds. Detailed rules are provided in the appendix.

Using the combination method, we generate 64 meta-action types, simulating the vehicle’s behavioral patterns in diverse scenarios. Each meta-action is represented as a continuous state, aligning with real-world human driving intentions.

\textbf{Behavior Justification Text Generation} The behavior justification text $\mathcal{T}$ provides a cause analysis of the ego vehicle's short-term driving strategy, enhancing the interpretability of the entire decision-making process. We utilize ChatGPT's API interface for generation: by using scene understanding QA annotations and meta-actions as contextual information, we prompt ChatGPT to automatically generate analytical justifications for the meta-actions. Compared to manual annotation methods, this approach produces more diverse behavior justification texts and can comprehensively and accurately reflect potential traffic factors (e.g., traffic rules) and social factors (e.g., social context)

\subsection{Training Pipeline}

To further enhance the perception-reasoning performance of Sce2DriveX, this paper introduces a task-oriented three-stage training pipeline, including: 1) Mixed Alignment Pre-training; 2) Scene Understanding Fine-tuning; and 3) End-to-End Driving Fine-tuning.

\textbf{Mixed Alignment Pre-training} This stage aligns the feature spaces of image/video representations and the LLM backbone. Sce2DriveX is pre-trained on CC3M image-text and WebVid-10M video-text datasets, covering diverse topics beyond autonomous driving. During this stage, the video encoder, image encoder, and LLM backbone weights are frozen, with only the shared projection parameters trained.

\textbf{Scene Understanding Fine-tuning} This stage enhances the model’s 3D spatial perception for hierarchical scene understanding. Sce2DriveX is fine-tuned on the dataset, using text cross-entropy loss to supervise its reasoning.

\textbf{End-to-End Driving Fine-tuning} This stage enhances the model’s long-axis task reasoning capabilities for interpretable end-to-end driving. Sce2DriveX is fine-tuned on the interpretable end-to-end driving dataset using the same training strategy as the scene understanding stage.

\begin{figure*}[!t]
\centering
\includegraphics[width=1\textwidth]{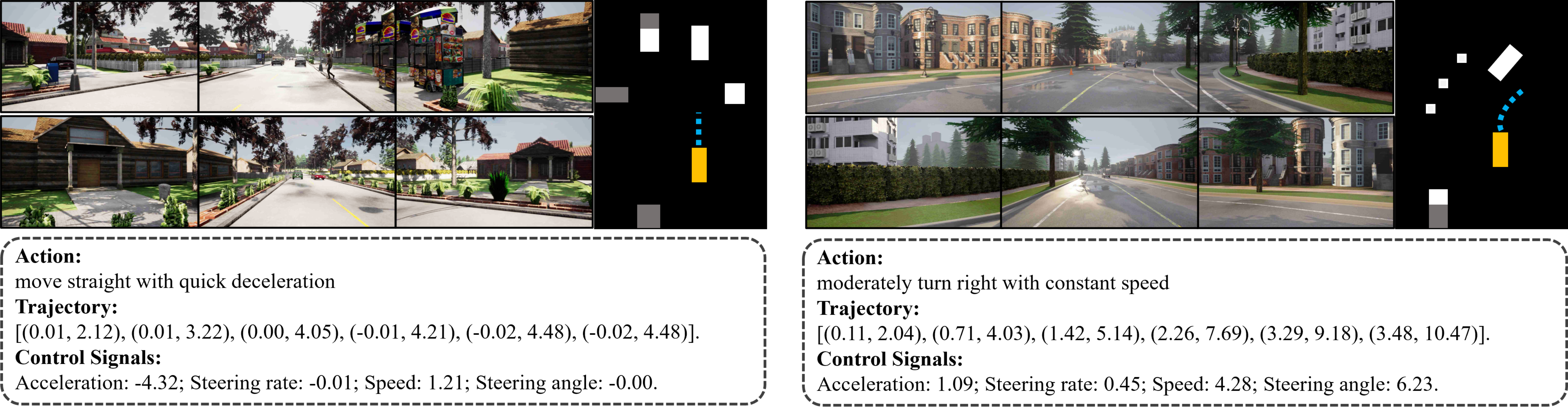} 
\caption{Visualization results of generalization testing (corner cases from the driving simulation dataset Bench2Drive).}
\label{fig4}
\vspace{-5pt} 
\end{figure*}

\section{Experiment}
\label{submission}

\subsection{Experimental Setup}

\textbf{Training Details:} We crop each image to a size of $224\times224$. We sample 8 frames evenly from each video and perform image preprocessing on each frame. Each batch of data includes a combination of images and videos. During the pre-training stage, the model is trained for 1 epoch with a batch size of 128, distributed across 6 A100 (80GB) GPUs. During the fine-tuning stage, AdamW optimizer and cosine learning rate scheduler are used, with an initial learning rate set to 2e-5, a warm-up ratio of 0.03, and a gradient accumulation step of 2. Specifically, scene understanding fine-tuning stage trains the model for 1 epoch, while end-to-end driving fine-tuning stage trains the model for 3 epochs. The entire process is completed on 8 L20 (48GB) GPUs, with a batch size of 4 per GPU.

\subsection{Hierarchical Scene Understanding}

We evaluate the performance of Sce2DriveX in hierarchical scene understanding task. Evaluation metrics include \textbf{NLP metrics}, such as \textbf{BLEU4 (B4)}, \textbf{ROUGE (R)}, \textbf{METEOR (M)}, \textbf{CIDEr (C)} and \textbf{accuracy (Acc)}. NLP metrics assess text generation quality, while accuracy measures consistency with GT labels. Notably, We focus only on the accuracy of traffic participant types. As no baselines exist for this task, Table \ref{tab1} reports only Sce2DriveX's results, aiming to inspire future research on hierarchical scene understanding in AD.

\subsection{Interpretable End-to-End Driving}

We report all evaluation metrics for interpretable end-to-end driving task. For motion planning (3s planned trajectory), the \textbf{L2 error ($m$)} is used. For meta-action reasoning, the \textbf{weighted accuracy $\alpha$Acc(\%)} is adopted, assigning certain weights to different parts of the meta-action, where the weight for steering estimation is 0.7 and the weight for speed estimation is 0.3. For behavior interpretation analysis, \textbf{NLP metrics} and \textbf{GPT scores} are utilized. For control signal generation, including the speed ($m/s$) and steering angle (\textdegree) for the next timestep, as well as the acceleration ($m/s^2$) and steering rate (\textdegree$/s$) for the current frame, the \textbf{Root Mean Square Error (RMSE)} is employed.

Table \ref{tab2} offers the comparison results between Sce2DriveX and baseline methods in motion planning (all experiments are conducted on the nuScenes dataset). Compared to small model-based methods such as NMP \cite{zeng2019end}, FF \cite{hu2021safe}, ST-P3 \cite{hu2022st}, UniAD \cite{hu2023planning}, and VAD \cite{jiang2023vad}, as well as LM-based methods like GPT-Driver \cite{mao2023gpt}, OmniDrive \cite{wang2024omnidrive}, and DriveVLM \cite{tian2024drivevlm}, Sce2DriveX achieves the lowest L2 error, demonstrating its effectiveness and robustness in planning human-like driving trajectories. It is worth noting that Sce2DriveX does not employ any implicit post-processing or data augmentation techniques to further enhance its performance.

To further validate the performance of Sce2DriveX in meta-action reasoning, behavior interpretation analysis, and control signal generation, we use the comprehensive VQA driving instruction dataset (Section 3.2) to implement the latest MLLM-based baseline methods: DriveGPT4 \cite{xu2024drivegpt4} and RAG-Driver \cite{yuan2024rag}. The parameter settings and model outputs remain consistent with the original work. Table \ref{tab3} compares Sce2DriveX with baselines, showing it achieves the highest accuracy, NLP metrics, and GPT scores, significantly enhancing end-to-end driving transparency and interpretability. Additionally, Sce2DriveX achieves lower RMSE for speed, steering angle, acceleration, and steering rate, demonstrating superior reasoning.

\begin{table}[!t]
\centering
\setlength{\tabcolsep}{5pt}  
\begin{tabular}{l|cccc|c}
\midrule
Element        & \multicolumn{1}{c}{M} & B4             & R              & C               & Acc            \\ \midrule
$E_{weather}$             & 60.89              & 70.13             & 87.75            & 713.68         
                          & 89.15  \\
$E_{road}$                & 75.70              & 77.07             & 87.64            & 757.63          
                          & 87.56  \\
$E_{facility}$            & 75.70              & 78.36             & 92.52            & 808.03              
                          & 93.71  \\
$E_{participant}$         & 58.06              & 63.03             & 76.97            & 621.21              
                          & 81.03  \\ \midrule 
\rowcolor{orange!25}
\textbf{Total}            & \textbf{62.16}     & \textbf{65.56}    & \textbf{78.41}   & \textbf{671.93}
                          & \textbf{85.69}  \\ \midrule
\end{tabular}
\caption{Quantitative results in hierarchical scene understanding.}
\label{tab1}
\end{table}

\begin{table}[!t]
\centering
\setlength{\tabcolsep}{10.35pt}  
\begin{tabular}{l|cccc}
\midrule
\multirow{2}{*}{Method} & \multicolumn{4}{c}{L2($m$)↓}                                     \\ 
                        & 1s            & 2s            & 3s            & Avg.          \\ \midrule
NMP                     & -             & -             & 2.31          & -             \\ 
ST-P3                   & 1.33          & 2.11          & 2.90          & 2.11          \\ 
FF                      & 0.55          & 1.20          & 2.54          & 1.43          \\ 
UniAD                   & 0.48          & 0.96          & 1.65          & 1.03          \\ 
VAD                     & 0.17          & 0.34          & 0.60          & 0.37          \\ \midrule
GPT-Driver              & 0.27          & 0.74          & 1.52          & 0.84          \\ 
OminiDrive              & 0.40          & 0.80          & 1.32          & 0.84          \\ 
DriveVLM                & 0.18          & 0.34          & 0.68          & 0.40          \\ \midrule
\rowcolor{orange!35}
\textbf{Sce2DriveX}     & \textbf{0.15} & \textbf{0.33} & \textbf{0.59} & \textbf{0.36} \\ \midrule
\end{tabular}
\caption{Comparison results in motion planning.}
\label{tab2}
\vspace{-5pt} 
\end{table}

\begin{table*}[!t]
\centering
\setlength{\tabcolsep}{5.2pt}  
\begin{tabular}{l|c|ccccc|cccc}
\midrule
\multirow{2}{*}{Method} & \multirow{2}{*}{$\alpha$Acc($\%$)↑ } & \multirow{2}{*}{M↑ } & \multirow{2}{*}{B4↑ } & \multirow{2}{*}{R↑ } & \multirow{2}{*}{C↑ } & \multirow{2}{*}{GPT↑ } & \multicolumn{4}{c}{RMSE}  \\
                     &                         &                   &                  &            
                     &                         &                   & acc($m/s^2$)↓                   
                     & rat(\textdegree$/s$)↓   & spd($m/s$)↓       & ang(\textdegree)↓  \\ \midrule
DriveGPT4            & 90.86                   & 18.75             & 5.18             & 31.89                                    & 130.60                  & 89.89             & 0.388            & 0.032           
                     & 0.096                   & 0.511   \\
RAG-Driver           & 93.06                   & 19.52             & 6.13             & 32.99                                    & 131.26                  & 90.78             & 0.271            & 0.025         
                     & 0.089                   & 0.449   \\ \midrule
\rowcolor{orange!35}
\textbf{Sce2DriveX}  & \textbf{94.29}          & \textbf{20.19}    & \textbf{6.69}    & \textbf{33.21}                           & \textbf{131.99}         & \textbf{91.11}    & \textbf{0.241}   & \textbf{0.021} 
                     & \textbf{0.081}          & \textbf{0.427}  \\ \midrule
\end{tabular}
\caption{Comparison results in meta-action reasoning, behavior justification analysis and control signal generation.}
\label{tab3}
\end{table*}

\begin{table*}[!t]
\centering
\setlength{\tabcolsep}{3.8pt}  
\begin{tabular}{l|c|ccccc|c|cccc}
\midrule
\multirow{2}{*}{Module Design} & \multirow{2}{*}{$\alpha$Acc($\%$)} & \multirow{2}{*}{M} & \multirow{2}{*}{B4} & \multirow{2}{*}{R} & \multirow{2}{*}{C} & \multirow{2}{*}{GPT} & \multirow{2}{*}{L2($m$)} & \multicolumn{4}{c}{RMSE}  \\
                     &                      &                      &                   &                 &                         &                      &                   & acc($m/s^2$)    & rat(\textdegree$/s$) & spd($m/s$)           & ang(\textdegree)  \\ \midrule
ICL                  & 81.46                & 12.69                & 2.11              & 23.45                                   & 101.36               & 75.56                & 1.01              & 0.401          
                     & 0.051                & 0.101                & 0.814  \\
w/o CoT              & -                    & -                    & -                 & -                                       & -                    & -                    & 0.38              & 0.246          
                     & 0.025                & 0.084                & 0.515  \\ 
w/o SU Fine-tuning   & 86.35                & 17.11                & 3.95              & 28.31                                   & 120.48               & 79.99                & 0.74              & 0.282          
                     & 0.038                & 0.089                & 0.675  \\
w/o Multi-view       & 91.56                & 19.17                & 5.95              & 31.74                                   & 124.29               & 83.14                & 0.51              & 0.250          
                     & 0.026                & 0.084                & 0.594  \\ 
w/o Image Encoder    & 94.00                & 19.19                & 6.28              & 31.95                                   & 126.76               & 86.25                & 0.39              & 0.243          
                     & 0.022                & 0.083                & 0.486  \\ \midrule
\rowcolor{orange!35}                     
\textbf{Baseline}    & \textbf{94.29}       & \textbf{20.19}       & \textbf{6.69}     & \textbf{33.21}                          & \textbf{131.99}      & \textbf{91.11}       & \textbf{0.36}     & \textbf{0.241} 
                     & \textbf{0.021}       & \textbf{0.081}       & \textbf{0.427}  \\ \midrule
\end{tabular}
\caption{Ablation study on the module design.}
\label{tab4}
\end{table*}

\subsection{Ablation Study}

We conduct ablation study to evaluate the contributions of each module in the Sce2DriveX framework. The specific settings include: 1) using only the in-context learning strategy to test the model (ICL); 2) excluding the QAs related to meta-action and justification from the dataset (w/o CoT); 3) omitting the scene understanding fine-tuning stage (w/o SU Fine-tuning); 4) using front-view scene videos as input (w/o Multi-view); and 5) removing the image encoder component  (w/o Image Encoder). Table \ref{tab4} offers the quantitative results of ablation study. We observe that the model performance declines when any module in the framework is altered or removed, thereby validating the rationality and effectiveness of the Sce2DriveX framework.

\subsection{Generalization Evaluation}

To further assess the generalization capability of Sce2DriveX, we conduct cross-dataset testing. Specifically, we obtain various styles of corner cases from the driving simulation dataset Bench2Drive \cite{jia2024bench2drive}. For concise visualization, only full-view images of the current frame are shown. As shown in Figure \ref{fig4}, we observe that Sce2DriveX is capable of generating satisfactory responses in a zero-shot transfer manner, demonstrating exceptional generalization.

\begin{figure}[!t]
\centering
\includegraphics[width=0.48\textwidth]{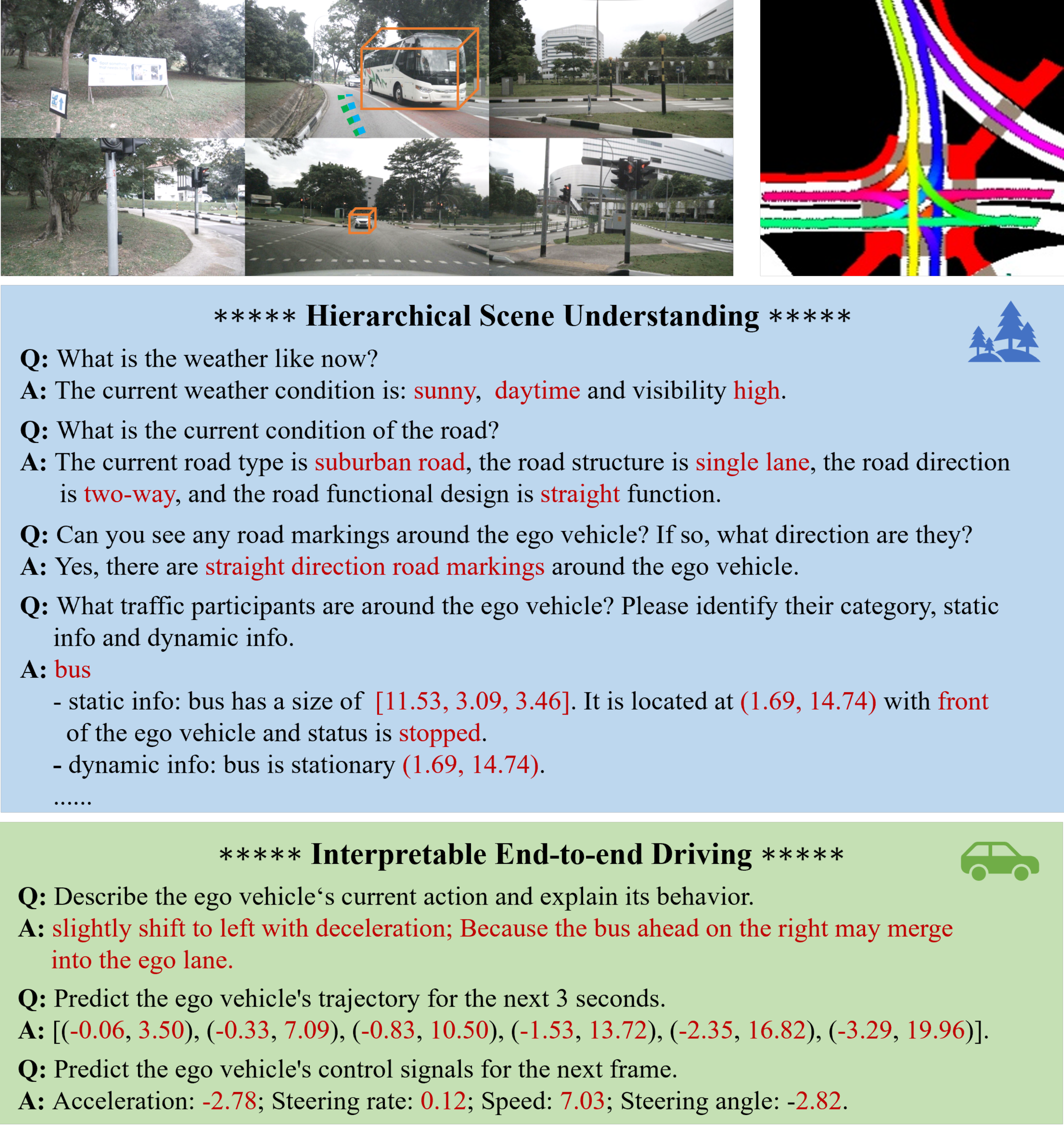} 
\caption{Qualitative demonstration of Sce2DriveX.}
\label{fig5}
\vspace{-5pt} 
\end{figure}

\subsection{Qualitative Demonstration}

Figure \ref{fig5} presents a visualized example of Sce2DriveX's reasoning in complex outdoor driving scenarios, showcasing its ability to perform progressive reasoning from hierarchical scene understanding to human-consensus end-to-end driving. The appendix provides additional qualitative comparisons and visual results of the  comprehensive VQA dataset.

\section{Conclusion}
\label{submission}

In this paper, we propose Sce2DriveX, enabling progressive reasoning from hierarchical scene understanding to interpretable end-to-end driving. Through multimodal learning of local scenes and global maps, it gains a deep understanding of long-range spatiotemporal relationships and road topology, enhancing generalization and consensus in cross-scene driving. We construct the first comprehensive VQA driving dataset for 3D spatial understanding and long-axis task reasoning and introduce a task-oriented three-stage supervised fine-tuning. Experimental results show Sce2DriveX excels in scene understanding, meta-action reasoning, behavior justification, motion planning, and control signal generation. We hope this work provides insights into MLLM applications in autonomous driving.

\nocite{langley00}

\bibliography{main}

\begin{thebibliography}{36}
\providecommand{\natexlab}[1]{#1}
\providecommand{\url}[1]{\texttt{#1}}
\expandafter\ifx\csname urlstyle\endcsname\relax
  \providecommand{\doi}[1]{doi: #1}\else
  \providecommand{\doi}{doi: \begingroup \urlstyle{rm}\Url}\fi

\bibitem[Alayrac et~al.(2022)Alayrac, Donahue, Luc, Miech, Barr, Hasson, Lenc, Mensch, Millican, Reynolds, et~al.]{alayrac2022flamingo}
Alayrac, J.-B., Donahue, J., Luc, P., Miech, A., Barr, I., Hasson, Y., Lenc, K., Mensch, A., Millican, K., Reynolds, M., et~al.
\newblock Flamingo: a visual language model for few-shot learning.
\newblock \emph{Advances in neural information processing systems}, 35:\penalty0 23716--23736, 2022.

\bibitem[Azuma et~al.(2022)Azuma, Miyanishi, Kurita, and Kawanabe]{azuma2022scanqa}
Azuma, D., Miyanishi, T., Kurita, S., and Kawanabe, M.
\newblock Scanqa: 3d question answering for spatial scene understanding.
\newblock In \emph{proceedings of the IEEE/CVF conference on computer vision and pattern recognition}, pp.\  19129--19139, 2022.

\bibitem[Caesar et~al.(2020)Caesar, Bankiti, Lang, Vora, Liong, Xu, Krishnan, Pan, Baldan, and Beijbom]{caesar2020nuscenes}
Caesar, H., Bankiti, V., Lang, A.~H., Vora, S., Liong, V.~E., Xu, Q., Krishnan, A., Pan, Y., Baldan, G., and Beijbom, O.
\newblock nuscenes: A multimodal dataset for autonomous driving.
\newblock In \emph{Proceedings of the IEEE/CVF conference on computer vision and pattern recognition}, pp.\  11621--11631, 2020.

\bibitem[Choudhary et~al.(2023)Choudhary, Dewangan, Chandhok, Priyadarshan, Jain, Singh, Srivastava, Jatavallabhula, and Krishna]{choudhary2023talk2bev}
Choudhary, T., Dewangan, V., Chandhok, S., Priyadarshan, S., Jain, A., Singh, A.~K., Srivastava, S., Jatavallabhula, K.~M., and Krishna, K.~M.
\newblock Talk2bev: Language-enhanced bird's-eye view maps for autonomous driving.
\newblock \emph{arXiv preprint arXiv:2310.02251}, 2023.

\bibitem[Das et~al.(2018)Das, Datta, Gkioxari, Lee, Parikh, and Batra]{das2018embodied}
Das, A., Datta, S., Gkioxari, G., Lee, S., Parikh, D., and Batra, D.
\newblock Embodied question answering.
\newblock In \emph{Proceedings of the IEEE conference on computer vision and pattern recognition}, pp.\  1--10, 2018.

\bibitem[Driess et~al.(2023)Driess, Xia, Sajjadi, Lynch, Chowdhery, Ichter, Wahid, Tompson, Vuong, Yu, et~al.]{driess2023palm}
Driess, D., Xia, F., Sajjadi, M.~S., Lynch, C., Chowdhery, A., Ichter, B., Wahid, A., Tompson, J., Vuong, Q., Yu, T., et~al.
\newblock Palm-e: An embodied multimodal language model.
\newblock \emph{arXiv preprint arXiv:2303.03378}, 2023.

\bibitem[Fukui et~al.(2016)Fukui, Park, Yang, Rohrbach, Darrell, and Rohrbach]{fukui2016multimodal}
Fukui, A., Park, D.~H., Yang, D., Rohrbach, A., Darrell, T., and Rohrbach, M.
\newblock Multimodal compact bilinear pooling for visual question answering and visual grounding.
\newblock \emph{arXiv preprint arXiv:1606.01847}, 2016.

\bibitem[Goyal et~al.(2017)Goyal, Khot, Summers-Stay, Batra, and Parikh]{goyal2017making}
Goyal, Y., Khot, T., Summers-Stay, D., Batra, D., and Parikh, D.
\newblock Making the v in vqa matter: Elevating the role of image understanding in visual question answering.
\newblock In \emph{Proceedings of the IEEE conference on computer vision and pattern recognition}, pp.\  6904--6913, 2017.

\bibitem[Hu et~al.(2021)Hu, Huang, Dolan, Held, and Ramanan]{hu2021safe}
Hu, P., Huang, A., Dolan, J., Held, D., and Ramanan, D.
\newblock Safe local motion planning with self-supervised freespace forecasting.
\newblock In \emph{Proceedings of the IEEE/CVF Conference on Computer Vision and Pattern Recognition}, pp.\  12732--12741, 2021.

\bibitem[Hu et~al.(2022)Hu, Chen, Wu, Li, Yan, and Tao]{hu2022st}
Hu, S., Chen, L., Wu, P., Li, H., Yan, J., and Tao, D.
\newblock St-p3: End-to-end vision-based autonomous driving via spatial-temporal feature learning.
\newblock In \emph{European Conference on Computer Vision}, pp.\  533--549. Springer, 2022.

\bibitem[Hu et~al.(2023)Hu, Yang, Chen, Li, Sima, Zhu, Chai, Du, Lin, Wang, et~al.]{hu2023planning}
Hu, Y., Yang, J., Chen, L., Li, K., Sima, C., Zhu, X., Chai, S., Du, S., Lin, T., Wang, W., et~al.
\newblock Planning-oriented autonomous driving.
\newblock In \emph{Proceedings of the IEEE/CVF Conference on Computer Vision and Pattern Recognition}, pp.\  17853--17862, 2023.

\bibitem[Jang et~al.(2017)Jang, Song, Yu, Kim, and Kim]{jang2017tgif}
Jang, Y., Song, Y., Yu, Y., Kim, Y., and Kim, G.
\newblock Tgif-qa: Toward spatio-temporal reasoning in visual question answering.
\newblock In \emph{Proceedings of the IEEE conference on computer vision and pattern recognition}, pp.\  2758--2766, 2017.

\bibitem[Jia et~al.(2024)Jia, Yang, Li, Zhang, and Yan]{jia2024bench2drive}
Jia, X., Yang, Z., Li, Q., Zhang, Z., and Yan, J.
\newblock Bench2drive: Towards multi-ability benchmarking of closed-loop end-to-end autonomous driving.
\newblock \emph{arXiv preprint arXiv:2406.03877}, 2024.

\bibitem[Jiang et~al.(2023)Jiang, Chen, Xu, Liao, Chen, Zhou, Zhang, Liu, Huang, and Wang]{jiang2023vad}
Jiang, B., Chen, S., Xu, Q., Liao, B., Chen, J., Zhou, H., Zhang, Q., Liu, W., Huang, C., and Wang, X.
\newblock Vad: Vectorized scene representation for efficient autonomous driving.
\newblock In \emph{Proceedings of the IEEE/CVF International Conference on Computer Vision}, pp.\  8340--8350, 2023.

\bibitem[Johnson et~al.(2017)Johnson, Hariharan, Van Der~Maaten, Fei-Fei, Lawrence~Zitnick, and Girshick]{johnson2017clevr}
Johnson, J., Hariharan, B., Van Der~Maaten, L., Fei-Fei, L., Lawrence~Zitnick, C., and Girshick, R.
\newblock Clevr: A diagnostic dataset for compositional language and elementary visual reasoning.
\newblock In \emph{Proceedings of the IEEE conference on computer vision and pattern recognition}, pp.\  2901--2910, 2017.

\bibitem[Lei et~al.(2018)Lei, Yu, Bansal, and Berg]{lei2018tvqa}
Lei, J., Yu, L., Bansal, M., and Berg, T.~L.
\newblock Tvqa: Localized, compositional video question answering.
\newblock \emph{arXiv preprint arXiv:1809.01696}, 2018.

\bibitem[Li et~al.(2023)Li, Li, Savarese, and Hoi]{li2023blip}
Li, J., Li, D., Savarese, S., and Hoi, S.
\newblock Blip-2: Bootstrapping language-image pre-training with frozen image encoders and large language models.
\newblock In \emph{International conference on machine learning}, pp.\  19730--19742. PMLR, 2023.

\bibitem[Lin et~al.(2023)Lin, Ye, Zhu, Cui, Ning, Jin, and Yuan]{lin2023video}
Lin, B., Ye, Y., Zhu, B., Cui, J., Ning, M., Jin, P., and Yuan, L.
\newblock Video-llava: Learning united visual representation by alignment before projection.
\newblock \emph{arXiv preprint arXiv:2311.10122}, 2023.

\bibitem[Liu et~al.(2024)Liu, Li, Li, and Lee]{liu2024improved}
Liu, H., Li, C., Li, Y., and Lee, Y.~J.
\newblock Improved baselines with visual instruction tuning.
\newblock In \emph{Proceedings of the IEEE/CVF Conference on Computer Vision and Pattern Recognition}, pp.\  26296--26306, 2024.

\bibitem[Ma et~al.(2022)Ma, Yong, Zheng, Li, Liang, Zhu, and Huang]{ma2022sqa3d}
Ma, X., Yong, S., Zheng, Z., Li, Q., Liang, Y., Zhu, S.-C., and Huang, S.
\newblock Sqa3d: Situated question answering in 3d scenes.
\newblock \emph{arXiv preprint arXiv:2210.07474}, 2022.

\bibitem[Mao et~al.(2023)Mao, Qian, Ye, Zhao, and Wang]{mao2023gpt}
Mao, J., Qian, Y., Ye, J., Zhao, H., and Wang, Y.
\newblock Gpt-driver: Learning to drive with gpt.
\newblock \emph{arXiv preprint arXiv:2310.01415}, 2023.

\bibitem[Qian et~al.(2024)Qian, Chen, Zhuo, Jiao, and Jiang]{qian2024nuscenes}
Qian, T., Chen, J., Zhuo, L., Jiao, Y., and Jiang, Y.-G.
\newblock Nuscenes-qa: A multi-modal visual question answering benchmark for autonomous driving scenario.
\newblock In \emph{Proceedings of the AAAI Conference on Artificial Intelligence}, volume~38, pp.\  4542--4550, 2024.

\bibitem[Sima et~al.(2025)Sima, Renz, Chitta, Chen, Zhang, Xie, Bei{\ss}wenger, Luo, Geiger, and Li]{sima2025drivelm}
Sima, C., Renz, K., Chitta, K., Chen, L., Zhang, H., Xie, C., Bei{\ss}wenger, J., Luo, P., Geiger, A., and Li, H.
\newblock Drivelm: Driving with graph visual question answering.
\newblock In \emph{European Conference on Computer Vision}, pp.\  256--274. Springer, 2025.

\bibitem[Tan \& Bansal(2019)Tan and Bansal]{tan2019lxmert}
Tan, H. and Bansal, M.
\newblock Lxmert: Learning cross-modality encoder representations from transformers.
\newblock \emph{arXiv preprint arXiv:1908.07490}, 2019.

\bibitem[Tian et~al.(2024)Tian, Gu, Li, Liu, Wang, Zhao, Zhan, Jia, Lang, and Zhao]{tian2024drivevlm}
Tian, X., Gu, J., Li, B., Liu, Y., Wang, Y., Zhao, Z., Zhan, K., Jia, P., Lang, X., and Zhao, H.
\newblock Drivevlm: The convergence of autonomous driving and large vision-language models.
\newblock \emph{arXiv preprint arXiv:2402.12289}, 2024.

\bibitem[Wang et~al.(2024)Wang, Yu, Jiang, Lan, Shi, Chang, Kautz, Li, and Alvarez]{wang2024omnidrive}
Wang, S., Yu, Z., Jiang, X., Lan, S., Shi, M., Chang, N., Kautz, J., Li, Y., and Alvarez, J.~M.
\newblock Omnidrive: A holistic llm-agent framework for autonomous driving with 3d perception, reasoning and planning.
\newblock \emph{arXiv preprint arXiv:2405.01533}, 2024.

\bibitem[Wu et~al.(2023)Wu, Han, Wang, Liu, Zhang, and Shen]{wu2023language}
Wu, D., Han, W., Wang, T., Liu, Y., Zhang, X., and Shen, J.
\newblock Language prompt for autonomous driving.
\newblock \emph{arXiv preprint arXiv:2309.04379}, 2023.

\bibitem[Xu et~al.(2024)Xu, Zhang, Xie, Zhao, Guo, Wong, Li, and Zhao]{xu2024drivegpt4}
Xu, Z., Zhang, Y., Xie, E., Zhao, Z., Guo, Y., Wong, K.-Y.~K., Li, Z., and Zhao, H.
\newblock Drivegpt4: Interpretable end-to-end autonomous driving via large language model.
\newblock \emph{IEEE Robotics and Automation Letters}, 2024.

\bibitem[Ye et~al.(2022)Ye, Huang, and Zeng]{ye2022visatlas}
Ye, Y., Huang, R., and Zeng, W.
\newblock Visatlas: An image-based exploration and query system for large visualization collections via neural image embedding.
\newblock \emph{IEEE Transactions on Visualization and Computer Graphics}, 2022.

\bibitem[Yu et~al.(2019)Yu, Xu, Yu, Yu, Zhao, Zhuang, and Tao]{yu2019activitynet}
Yu, Z., Xu, D., Yu, J., Yu, T., Zhao, Z., Zhuang, Y., and Tao, D.
\newblock Activitynet-qa: A dataset for understanding complex web videos via question answering.
\newblock In \emph{Proceedings of the AAAI Conference on Artificial Intelligence}, volume~33, pp.\  9127--9134, 2019.

\bibitem[Yuan et~al.(2024)Yuan, Sun, Omeiza, Zhao, Newman, Kunze, and Gadd]{yuan2024rag}
Yuan, J., Sun, S., Omeiza, D., Zhao, B., Newman, P., Kunze, L., and Gadd, M.
\newblock Rag-driver: Generalisable driving explanations with retrieval-augmented in-context learning in multi-modal large language model.
\newblock \emph{arXiv preprint arXiv:2402.10828}, 2024.

\bibitem[Zeng et~al.(2019)Zeng, Luo, Suo, Sadat, Yang, Casas, and Urtasun]{zeng2019end}
Zeng, W., Luo, W., Suo, S., Sadat, A., Yang, B., Casas, S., and Urtasun, R.
\newblock End-to-end interpretable neural motion planner.
\newblock In \emph{Proceedings of the IEEE/CVF Conference on Computer Vision and Pattern Recognition}, pp.\  8660--8669, 2019.

\bibitem[Zhang et~al.(2023)Zhang, Li, and Bing]{zhang2023video}
Zhang, H., Li, X., and Bing, L.
\newblock Video-llama: An instruction-tuned audio-visual language model for video understanding.
\newblock \emph{arXiv preprint arXiv:2306.02858}, 2023.

\bibitem[Zhang et~al.(2021)Zhang, Li, Hu, Yang, Zhang, Wang, Choi, and Gao]{zhang2021vinvl}
Zhang, P., Li, X., Hu, X., Yang, J., Zhang, L., Wang, L., Choi, Y., and Gao, J.
\newblock Vinvl: Revisiting visual representations in vision-language models.
\newblock In \emph{Proceedings of the IEEE/CVF conference on computer vision and pattern recognition}, pp.\  5579--5588, 2021.

\bibitem[Zhu et~al.(2023{\natexlab{a}})Zhu, Lin, Ning, Yan, Cui, Wang, Pang, Jiang, Zhang, Li, et~al.]{zhu2023languagebind}
Zhu, B., Lin, B., Ning, M., Yan, Y., Cui, J., Wang, H., Pang, Y., Jiang, W., Zhang, J., Li, Z., et~al.
\newblock Languagebind: Extending video-language pretraining to n-modality by language-based semantic alignment.
\newblock \emph{arXiv preprint arXiv:2310.01852}, 2023{\natexlab{a}}.

\bibitem[Zhu et~al.(2023{\natexlab{b}})Zhu, Chen, Shen, Li, and Elhoseiny]{zhu2023minigpt}
Zhu, D., Chen, J., Shen, X., Li, X., and Elhoseiny, M.
\newblock Minigpt-4: Enhancing vision-language understanding with advanced large language models.
\newblock \emph{arXiv preprint arXiv:2304.10592}, 2023{\natexlab{b}}.

\end{thebibliography}
\bibliographystyle{icml2025}


\end{document}